\documentclass[10pt,twocolumn,letterpaper]{article}
\usepackage[ruled]{algorithm2e}
\usepackage{indentfirst}
\usepackage{cvpr}              

\title{LanePtrNet: Revisiting Lane Detection as Point Voting and Grouping on Curves}
\begin{document}
\author{
Jiayan Cao\footnotemark[1], Xueyu Zhu\footnotemark[1], Cheng Qian\footnotemark[2]\\
Appen\\
Fudan University\\
{\tt\small jcao2@appen.com,22210720334@m.fudan.edu.cn,cqian@appen.com}
}

\maketitle
\renewcommand{\thefootnote}{\fnsymbol{footnote}} 
\footnotetext[1]{These authors contributed equally to this work.} 
\footnotetext[2]{Corresponding authors.} 
\begin{abstract}
Lane detection plays a critical role in the field of autonomous driving. Prevailing methods generally adopt basic concepts (anchors, key points, etc.) from object detection and segmentation tasks, while these approaches require manual adjustments for curved objects, involve exhaustive searches on predefined anchors, require complex post-processing steps, and may lack flexibility when applied to real-world scenarios.In this paper, we propose a novel approach, LanePtrNet, which treats lane detection as a process of point voting and grouping on ordered sets: Our method takes backbone features as input and predicts a curve-aware centerness, which represents each lane as a point and assigns the most probable center point to it. A novel point sampling method is proposed to generate a set of candidate points based on the votes received. By leveraging features from local neighborhoods, and cross-instance attention score, we design a grouping module that further performs lane-wise clustering between neighboring and seeding points. Furthermore, our method can accommodate a point-based framework, (PointNet++ series, etc.) as an alternative to the backbone. This flexibility enables effortless extension to 3D lane detection tasks. We conduct comprehensive experiments to validate the effectiveness of our proposed approach, demonstrating its superior performance.

\end{abstract}    
\section{Introduction}
\label{sec:intro}
Lane detection \cite{Alpher01} in driving scenes is a fundamental issue in the functionality of autonomous vehicles. This module encompasses the identification and tracking of lane markings, ensuring that vehicles remain within their designated lanes. The precise detection of lane lines is of utmost importance for various operations such as lane-keeping, lane changes, and adaptive cruise control, all of which contribute to the advancement of automation in driving.\\
\begin{figure}
    \centering
    \includegraphics[width=1\linewidth]{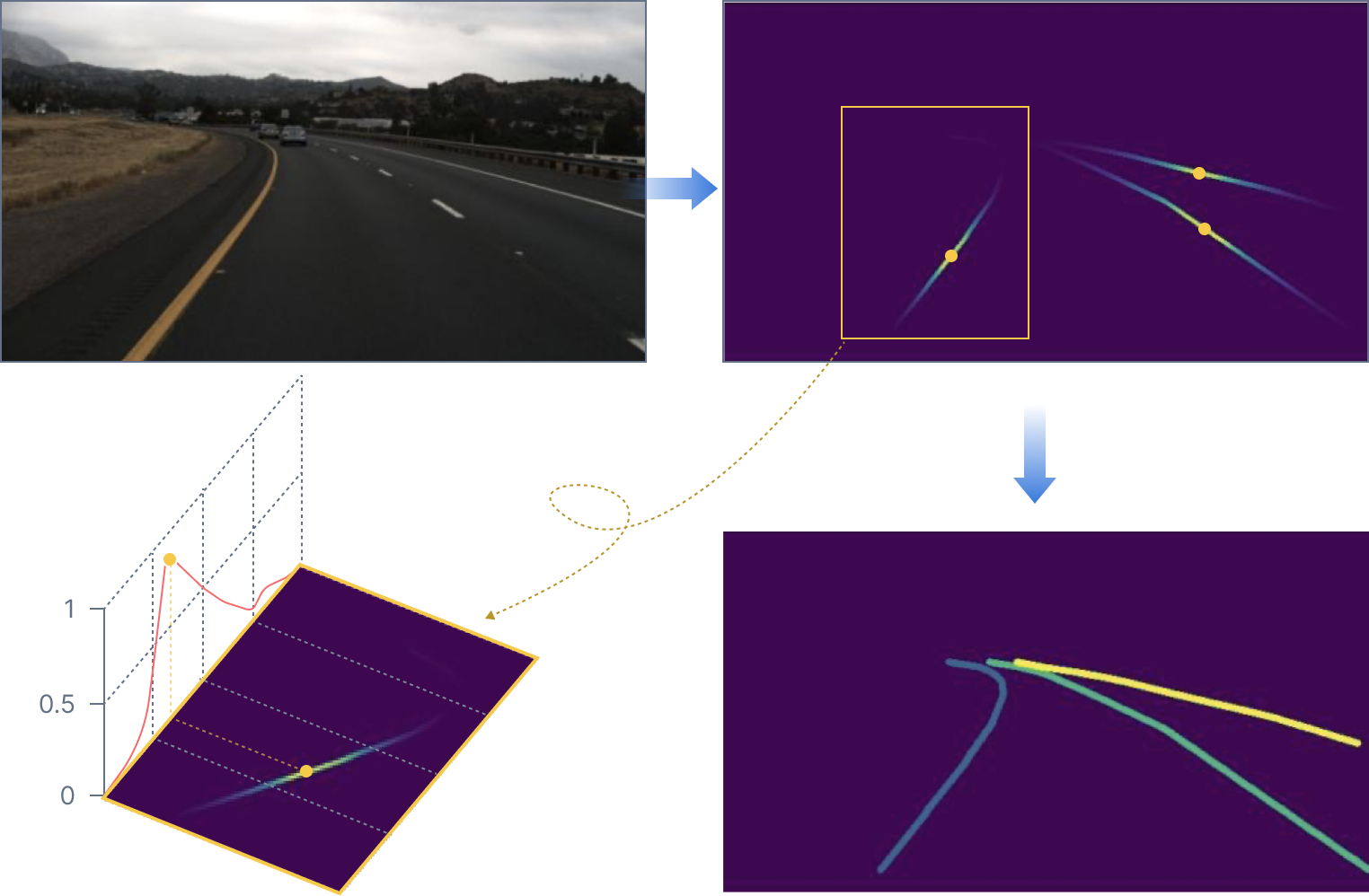}
    \caption{Proposed method introduces the concept of centerness as a fundamental scoring metric for candidate proposals in lane detection. Points located along the lane are represented as a bell-shaped distribution, with the curve centers ignited at the peak denoted by yellow dots. This formulation enables the reformation of lane detection as a grouping task based on the provided seeds.}
    \label{fig:inro}
\end{figure}
In the realm of academic research, lane detection techniques leveraging deep learning can be classified into four distinct categories: segmentation-based, anchor-based, parameter-curve-based, and keypoint-based. Segmentation-based methods \cite{Alpher03,Alpher04,Alpher05,Alpher15,Alpher16} involve the prediction of individual foreground points in the input image at a per-pixel level, with each class representing a specific type of lane line. However, these approaches encounter inherent challenges, notably the presence of a dynamic number of lane lines. Addressing this challenge requires the incorporation of an upstream detection task to produce instance segmentation results. Furthermore, the conversion of the segmentation mask output into a vector representation of lane lines necessitates the implementation of complex post-processing procedures. While Anchor-based methods \cite{Alpher06,Alpher07,Alpher08,Alpher09,Alpher19} necessitates the prior specification of a predetermined set of linear anchors. These anchors serve as reference points for the regression of offset values between sampled points and the established anchors, thereby facilitating the detection of lane lines. Although this approach has demonstrated promising outcomes, its adaptability is limited by the diverse range of lane line shapes. Additionally, this method requires careful design of anchor settings and the assignment of ground truth to these predefined elements. Considering the curved nature commonly observed in lane objects, parameter-curve based methods \cite{Alpher10,Alpher11,Alpher12} adopt higher-order polynomials or Bézier splines to fit the lane lines. This approach results in remarkably smooth lane line shapes and great flexibility in re-sampling control points on curves. However, it is crucial to ensure that the original datasets are appropriately organized in the corresponding format for these methods. Improper transformation from annotations, such as key-points or mask results, can lead to either under-fitting or over-fitting of the curves, consequently impairing the overall performance of the lane detection system.

To mitigate the limitations associated with the aforementioned approaches, we present LanePtrNet, a novel model that addresses these challenges. LanePtrNet predicts multiple key points along the lane markings and utilizes these points to infer the overall shape of the lane. This approach eliminates the requirement for intricate post-processing steps and enables efficient inference speeds. Notably, LanePtrNet exhibits similarities with keypoint-based methods \cite{Alpher13,Alpher14,Alpher17,Alpher02}, aligning with their shared characteristics and advantages, achieves state-of-the-art performance on TuSimple \cite{Alpher20} and CULane \cite{Alpher03} dataset. Key contributions of our work can be summarized as follows:
\begin{itemize}
  \item Our study introduces a novel framework that redefines lane detection as a process of seed point voting and grouping on ordered sets. This departure from conventional methodologies, which primarily rely on principles from object detection and segmentation tasks, distinguishes our proposed framework.
  \item  We introduce the concept of centerness on the curves as a fundamental aspect of lane detection. To predict high-quality candidates, we develop a center-aware sampling method that effectively incorporates this concept.
  \item  We propose a straightforward yet highly efficient grouping module that capitalizes on local neighborhood features and global semantic information. This module facilitates lane-wise clustering between neighboring and seeding points, thereby enabling LanePtrNet to handle various real-world scenarios.
\end{itemize}

\section{Related Work}
\noindent
\textbf{Segmentation-based Methods} Early research \cite{Alpher03}  \cite{Alpher04} on lane marking detection primarily followed the semantic segmentation paradigm\cite{fcn_seg}\cite{chen2017deeplab}, where each lane was treated as a distinct semantic class and individual pixels were assigned to their corresponding lane category.  However, in complex road situations involving intersections, turns, and intricate traffic patterns, the number of detected lane lines can vary significantly, rendering a fixed size of output semantic classes unsuitable. Consequently, many recent efforts in lane detection have shifted their focus towards instance segmentation\cite{maskrcnn}\cite{rfcn}, which allows for more precise lane information. MMANet \cite{Alpher05} employs attention mechanisms to aggregate local and global memory information at different scales and then integrate with the features of the current frame to obtain instance segmentation results for lane lines. CondLaneNet \cite{Alpher15} predicts the starting point of each lane line and utilizes conditional convolution to generate instances of each lane line. Chen \textit{et al.} \cite{Alpher16} employ transformers to generate dynamic kernels for each lane, enhancing global structure understanding.\\
\textbf{Anchor-based Methods} These methods associate predefined geometric shapes (anchor boxes, linear segments, curvatures \etc) with each pixel and detect lane lines by regressing the relative deformation offsets. Some approaches\cite{Alpher06} \cite{Alpher08} incorporate different feature extractors to enhance their performance. Certain studies emphasize the importance of measuring the similarity between predefined shapes and ground-truth lane markings. CLRNet \cite{Alpher09} improves lane detection by introducing the Line IoU (Intersection over Union) loss to enhance accuracy. Based on Line IoU, CLRerNet \cite{Alpher07} introduces Lane IoU as a learnable parameter enhancing the flexibility of adaptive thresholding. In order to overcome the rigidity of predefined anchor boxes, ADNet \cite{Alpher19} decomposes them into a heatmap for learning the starting points and their associated directions. 

In essence, anchor-based methods rely on the utilization of pre-defined anchor boxes. However, the intricate and diverse shapes exhibited by lane markings pose a significant challenge in capturing these complex patterns. As the input resolution increases, the number of possible lane combinations grows exponentially, given that lanes can exist in arbitrary lengths and directions. Defining a perfect geometry becomes uncertain in this context, and the underlying feature extraction process based on these anchor settings becomes a formidable task, \eg accurately predicting curved and dotted lines based on rotated boxes, presents additional complexities\cite{orcnn}\cite{li2021fcosr}. In comparison, our method exhibits a high degree of flexibility which can handle any arbitrary lane types as it treats each pixel as an individual point. \\
\textbf{Parameter-based Methods} Rather than relying on the definition of anchors, these methods adopt an end-to-end learning approach to directly learn the curve parameters of each lane. Feng \textit{et al. }\cite{Alpher22} proposes an architecture consisting of a deep network that predicts a lane segmentation map and a differentiable least-squares fitting module that optimizes the parameters of the best-fitted curve for each map. PolyLaneNet \cite{Alpher10} employs fully connected layers to directly estimate the polynomial coefficients for lane fitting while BézierLaneNet \cite{Alpher12} predicts the relevant parameters of curves (i.e. control points or Bézier coefficients) to achieve lane forecasting. 

Although aimed to predict smoothed curves, these methods may encounter difficulties in accurately learning the exact parameters when the data formats are not initially prepared in their forms, \eg datasets that are annotated by humans often consist of arbitrarily placed lane points rather than sets of polynomial coefficients. Consequently, these methods require a dedicated pre-processing step to formulate the parameters as ground-truth signals. The capability of the underlying model largely depends on the quality of this fitting process where any instances of under-fitting or over-fitting in the data samples can lead to a significant decrease in the overall system's performance. \\
\textbf{Keypoint-based Methods} Recent works view lane line detection as a problem of predicting key points and regressing offset values\cite{pose_rsn}\cite{moon2019posefix}. FOLOLane \cite{Alpher02} outputs a heatmap to determine whether a pixel is a keypoint and predicts offsets to accurately compensate for the keypoint's position. PINet \cite{Alpher17} employs a stacked hourglass network for predicting keypoint positions and feature embeddings, which are subsequently used to cluster lane instances based on their similarity. GANet \cite{Alpher14} involves associating predicted key points with their respective lane lines by globally predicting their offset to the starting point of the corresponding lane. RCLane \cite{Alpher13} utilizes a relay chain to represent lane markings, predicting key points and their consecutive relationships in the lane using the distance head and transfer head. These methods usually require intricate post-processing steps, \eg Non-Maximum Suppression (NMS). In contrast, our approach significantly accelerates this procedure and can be executed in an end-to-end fashion by employing centerness sampling techniques to obtain candidate lane points and then utilizing a cross-instance attention module to generate the final lane lines. 
\section{Method}
\subsection{Overall Architecture}
\begin{figure*}
    \centering
    \includegraphics[width=0.92\linewidth]{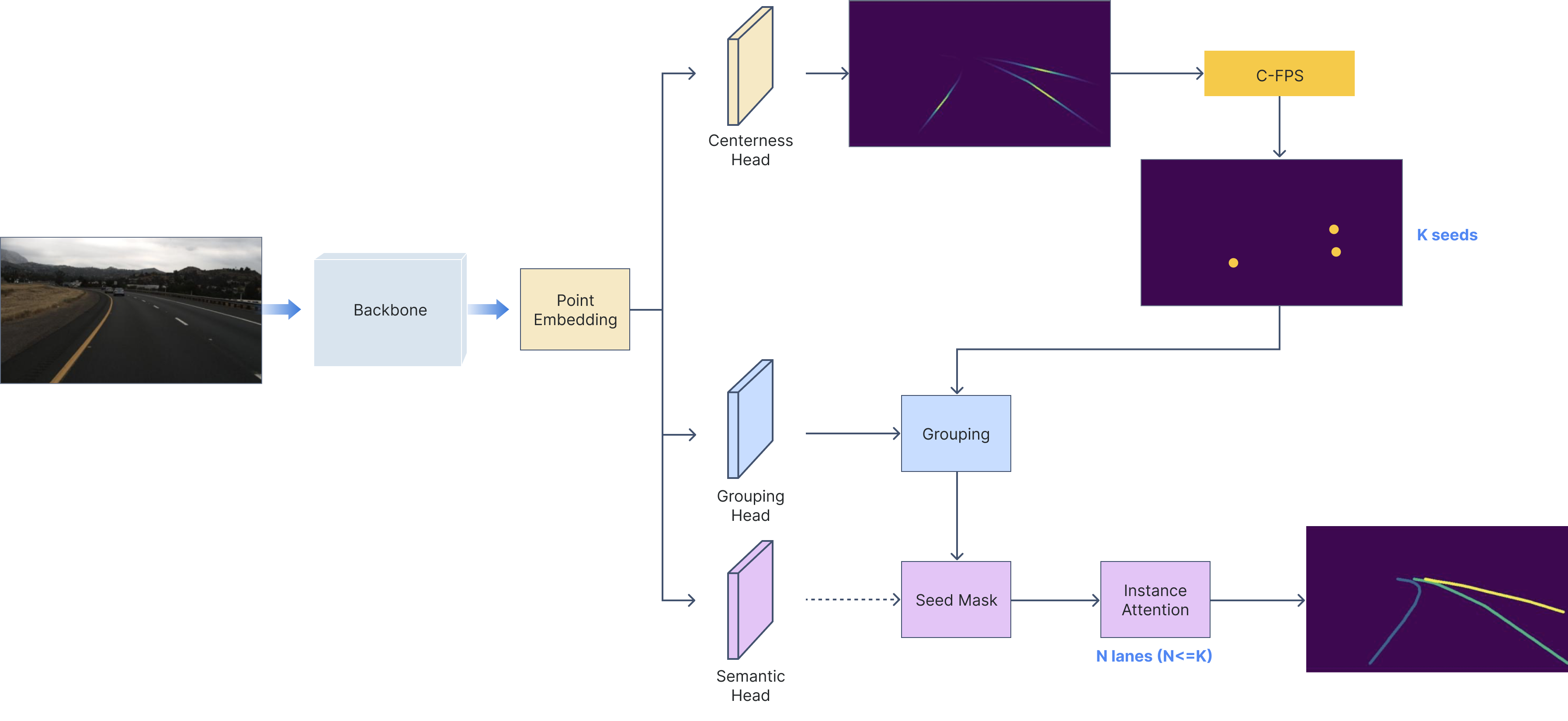}
    \caption{The overall architecture of LanePtrNet. Given an input image from the front view, the backbone encoder is employed to extract point-wise features. In the decoding phase, the centerness head is employed to predict the centered distribution of lane lines, generating a confidence map. Subsequently, the C-FPS algorithm is applied to identify $K$ seed points for the lanes. The grouping head performs clustering based on the positions of these seed central points, producing seed masks. Following this, the cross-instance attention module determines $N$ (where $N\leq{K}$) lane lines as the final prediction results. Additionally, the semantic head provides global semantic information about the lane lines to guide the generation of seed masks.}
    \label{fig:a}
\end{figure*}
The overall architecture of LanePtrNet, as depicted in \cref{fig:a}, follows an encode-decoder paradigm. Each pixel within the input image is regarded as a point within a set, characterized by spatial coordinates and RGB channels serving as features. The encoder component employs a convolutional backbone (\eg \cite{resnet}\cite{Alpher23}), which takes the raw input image and generates pixel-wise embeddings. Alternatively, a point-based backbone \cite{pointnet}\cite{Alpher25}\cite{voxelnet}\cite{minkunet}\cite{spvcnn}\cite{pvrcnn} can also be employed for a purely point-based encoder.

The decoder module consists of multiple heads that receive the point embeddings and predict the final lane results. Specifically, the decoder section comprises two key components: the centerness head and the grouping head. The centerness head is responsible for predicting the peak response for the centered distribution of lane lines. These points are then utilized in conjunction with the Centerness Farthest Point Sampling (C-FPS) algorithm to identify seed points.

In order to determine which pixel points belong to the selected seed candidates and form the final lane results, a binary classification task is employed to distinguish between foreground and background (\ie points on the lane or not). Points surpassing a certain classification threshold, given their corresponding seed points, are considered to be part of the lane.
Following this, the grouping head combines features from both the seed points and pixel embeddings, resulting in colored outputs for each input seed. Inspired by the definition of the dice coefficient, an instance-aware attention module is utilized to eliminate duplicate lanes, thereby producing a more concise output.

During the training stage, a semantic head is incorporated as an auxiliary component within the decoder section. This enhances the network's ability to acquire a comprehensive understanding of global information, enabling it to determine whether current pixels are associated with lane lines.
\subsection{Curved Centerness Prediction}
In object detection tasks\cite{centernet}\cite{Alpher22}\cite{gfocal_loss}\cite{gfocal_loss_v2}, IoU or centerness are two popular metrics that measure the quality of predicted objects. Typical approaches including anchor-free or anchor-based methods involve associating each pixel or anchor with the ground-truth objects through a label assignment process. This process determines an optimal allocation of positive, negative, and ignored samples, thereby providing a score that quantifies the suitability or quality of the resulting match, contributing to better performance. 

To be specific, to find the center point of a target, baseline approaches \cite{Alpher22}\cite{centernet} introduce the concept of "centerness". Given the distances of a pixel to the top, bottom, left, and right boundaries of a bounding box, denoted as $t$ (top), $b$ (bottom), $l$ (left), and $r$ (right), centerness is defined as follows:\\
\begin{equation}
centerness=\sqrt{\frac{min(l,r)}{max(l,r)}*\frac{min(t,b)}{max(t,b)}}
\label{eq:fcos}
\end{equation}

\begin{figure}
    \centering
    \includegraphics[width=0.5\linewidth]{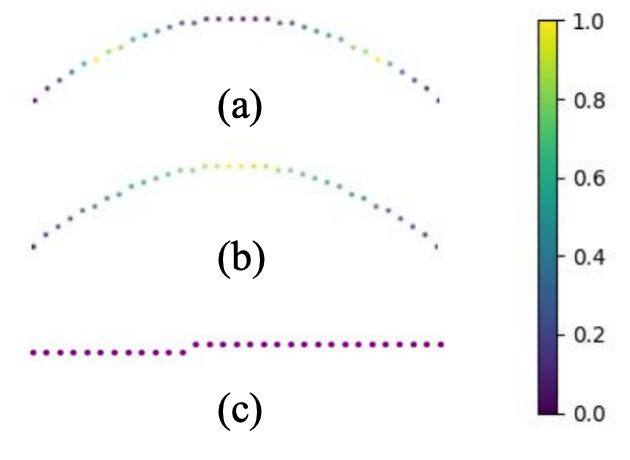}
    \caption{Computing the centerness heatmaps of curves where points in yellow depict peak value. (a) Using \cref{eq:fcos}. (b) Using the path integral method \cref{eq:changeSi}. (c) Corner cases where lane markings approximate extreme aspect ratio may lead to unstable box centerness (\eg a straight line).}
    \label{fig:boxcurve}
\end{figure}

However, this approach can not be directly adopted for curved objects, as it has two main drawbacks. The first one is the tendency to encounter multi-modal issues, as illustrated in \cref{fig:boxcurve} (a) when computing curved lane lines. The second one arises when the lane lines happen to be perfectly straight and aligned horizontally, as depicted in \cref{fig:boxcurve} (c). In such cases, this method fails to identify the center point of the lane lines. To overcome the aforementioned drawbacks, we propose a curve-aware centerness that computes the accumulated arc length distances between the connecting points. To start with, a typical mathematical formulation of the line integral along a curve \begin{math} \mathcal{C} \end{math} is defined as:
\begin{equation}
\int\limits_{C}^{}f(x,y)ds=\int\limits_{a}^{b}f(r(t))r'(t)dt
\label{eq:arc}
\end{equation}
where \begin{math} \mathbf{r} \colon [a,b]\to \mathcal{C} \end{math} is an arbitrary bijective parametrization of the curve \begin{math} \mathcal{C} \end{math} such that \begin{math} r(a)\end{math} and \begin{math} r(b)\end{math} give the endpoints of \begin{math} \mathcal{C} \end{math} and \begin{math} a < b\end{math}. Taking into account the discrete nature of image data, assuming a road lane has $M$ key points, with the starting point at $x_{1}$ and the ending point at $x{M}$, the total arc length of the $k$-th point on this lane can be calculated on the ordered set using the following formula:
\begin{equation}
S_{k}=\displaystyle\sum_{i=1}^{k}{|{x}_{i}-{x}_{i-1}|}^{2}
\label{eq:path}
\end{equation}

Assuming the central points of the curve give the highest centerness score and $S_{center}=0.5$, we transform the accumulated distance of this ordered set into a mono-modal distribution where centerness is normalized between 0 and 1, 0 for the starting and ending points and 1 for the most centered points.
Calculating the integral of the arc length from each key point to the starting point, and then dividing it by the total arc length, yields a relative value $S_{i}$ for each key point. Clearly, $S_{i}$ increases monotonically from the starting point to the endpoint, ranging between 0 and 1. We consider the keypoint at the position where $S_{i}$ takes the value of 0.5 as the central point. Using \cref{eq:changeSi}, the $C_{i}$ value corresponding to the central point is set to 1, and the $C_{i}$ values corresponding to points farther from the center decrease.
\begin{equation}
{C}_{i}=1-\frac{|{S}_{i}-0.5|}{0.5}
\label{eq:changeSi}
\end{equation}
As shown in \cref{fig:boxcurve} (b), the arc-length path integration method proves to be effective in mitigating the issue of multiple peaks observed in \cref{fig:boxcurve} (a). Furthermore, this approach allows for the effortless calculation of the center point of the lane lines in scenarios such as the one depicted in \cref{fig:boxcurve} (c).
The center head predicts a centerness score for each pixel and we extend the original algorithms \cite{Alpher25}\cite{sasa-fps}\cite{ic-fps}to Center-Farthest Point Sampling (C-FPS) algorithm to vote the center points on the curves.
\begin{algorithm}
\caption{Centerness-Farthest Point Sampling}\label{alg:three}
\KwData{point coordinates $X=\{x_{1}$, ..., $x_{N}\}$, with corresponding centerness scores $C=\{c_{1}$, ..., $c_{N}\}$, $N$ is the number of points}
\KwResult{sampled set of points $K$=$\{k_{1}$, ..., $k_{M}\}$}
Initialize sampled indices $K = \{argmax(C)\}$\;
Initialize a distance array $D$ of length $N$ with $+\inf$\;
\For {$i$ = 1 to $M$} {
    $p = K[-1]$\;
    \For {$j = 1$; $j \leq N $ \&\& $j$ not in $K$; $j+=1$} 
    {
    $d_{j}$ = $c_{j}^\gamma * \| x_{j} - x_{p}\|$
    }
    $k_{i}$ = argmax($D$)\;
    $K$ = $K \cup \{k_{i}\}$\;
}
\For {$i$ = 1 to $M$} {
    $p = K[-1]$\;
    \For {$j = 1$; $j \leq N $ \&\& $j$ not in $K$; $j+=1$} 
    {
    $d_{j}$ = $c_{j}^\gamma * \| x_{j} - x_{p}\|$
    }
    $k_{i}$ = argmax($D$)\;
    $K$ = $K \cup \{k_{i}\}$\;
}
\end{algorithm}
The hyper-parameter \(\gamma\) is utilized to regulate the weighted significance of centerness and spatial distance. A higher value emphasizes locally high-confident peaks, whereas a lower value promotes the presence of diverse lane centers across the entire image. As shown in \cref{fig:boxcurve}, the C-FPS algorithm is employed to identify the center points of all lane lines within an image, utilizing the confidence scores derived from the centerness map and their corresponding coordinates. Given that the image is represented in two dimensions and the original FPS strategy operates on 3D data (\ie $xyz$ coordinates), we restrict the input on the z-axis to solely compute the distance across $xy$ panels.
It is noteworthy that there is a potential for enhancement by integrating depth prediction as the input for the z-axis. This is particularly relevant in the context of lane lines, initially represented in 3D world coordinates and subsequently projected onto the camera perspective when given calibration information.
\subsection{Lane Points Grouping}
\begin{figure}
    \centering
    \includegraphics[width=0.9\linewidth]{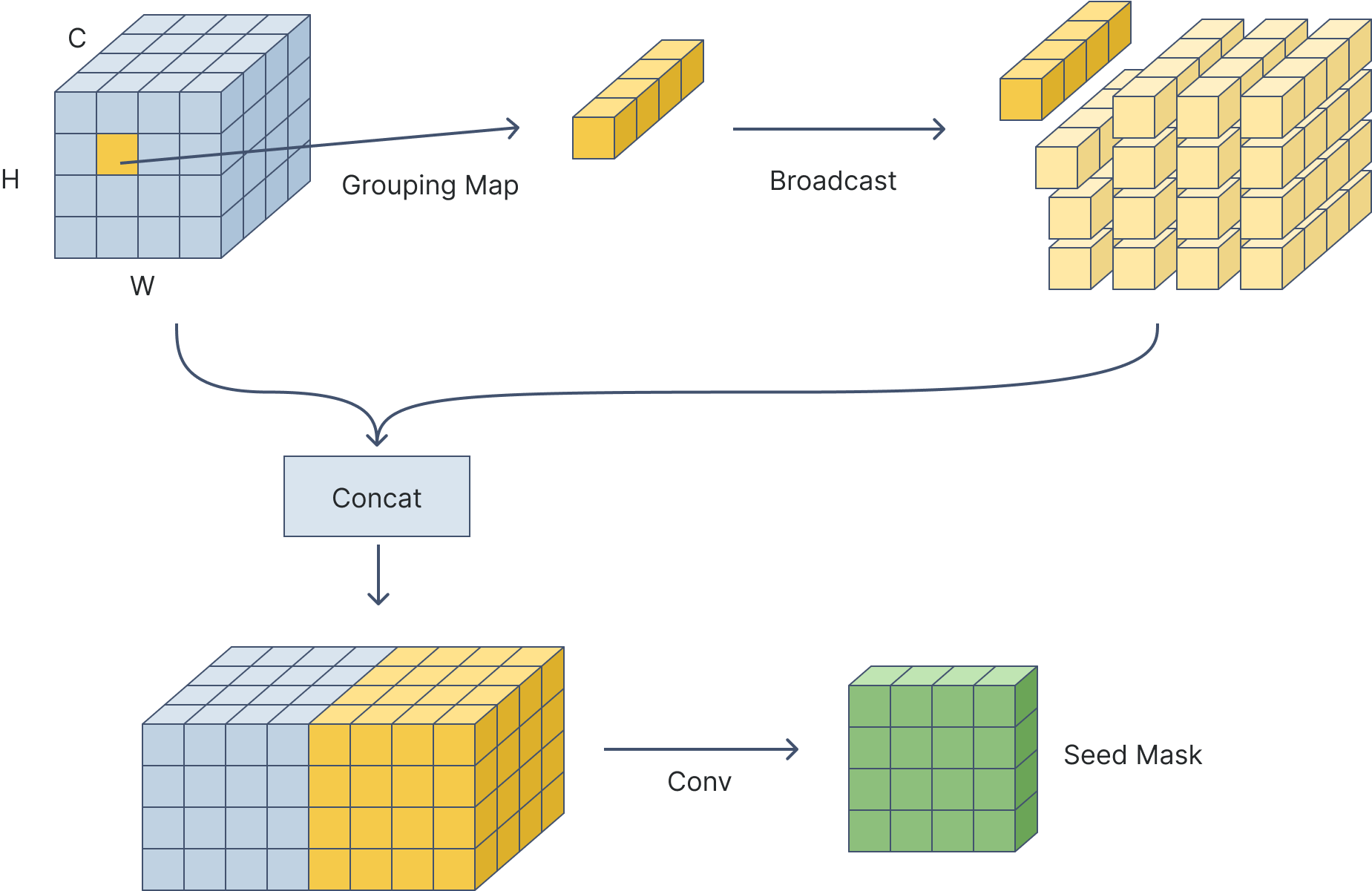}
    \caption{The process of generating lane groups. Blue blocks represent the grouping map while the yellow cubes are seed points aggregating their corresponding features. Through tensor broadcasting, these selected seeds are duplicated and concatenated to match the size of the spatial size, obtaining the final seed mask after several convolutional blocks.}
    \label{fig:seek mask}
\end{figure}

As shown in \cref{fig:seek mask}, the model's group head produces a control map with dimensions of $C\times{H}\times{W}$. We select the seed features according to their spatial coordinates in this feature map. These selected features are repeated to match the size of $H\times{W}$ and then concatenated with the group map, resulting in a feature map with dimensions of $2C$ with the same spatial size. In the last stage, a series of convolutional blocks with batch normalization and rectified linear unit (ReLU) activation function are employed to generate the final seed mask. The seed points are selected and indexed in a top-down manner, while the control map serves as a global feature in a bottom-up fashion. This lightweight head efficiently implements seed features through tensor repeating and broadcasting techniques, minimizing the computational footprint on the GPU memory.
\subsection {Cross-Instance Attention Voting}
 As the standard post-processing strategy (\ie NMS) is designed to discard entities that are below a given probability bound (confidence, quality scores, \etc), We choose centerness confidence and cross-instance attention scores as the key factors to filter out unique entities out of duplicate proposals.
Assume there are two instance groups, containing seed points and their corresponding clustered lane points, the standard Intersection over Union on sets as follows:
\begin{equation}
IoU_{X,Y} =\frac{2\left | X\cap Y\right |}{\left | X\right |+\left | Y\right |}
\label{eq:set_iou}
\end{equation}
where $X$ represents the number of lane line pixels in the predicted seed mask, and $Y$ represents the number of lane line pixels in the ground truth.
As the output of the grouping head is probability distribution indicating whether this point belongs to the current seed, we modify it to suit continuous input as follows:
\begin{equation}
\hat{IoU}_{X,Y} =\frac{2X\times Y}{X^2 + Y^2}
\label{eq:set_soft_iou}
\end{equation}
Here $\times$ stands for element-wise multiplication. Assume we have $K$ seed groups, a $K\times{K}$ attention map is generated, where each attention score between the $i$-th and $j$-th lane is computed as $a_{i,j} = IoU(X_{i},X_{y})$. During the inference phase, the seed set is rearranged and searched based on their centerness confidence by descending order. Subsequently, the remaining seeds that exceed the attention score threshold are eliminated, resulting in the final $N$ lanes where $N \leq K$.

This scoring mechanism offers several advantages. Firstly, it exhibits computational efficiency can be optimized on GPU devices, as the input size is pre-filtered using C-FPS to avoid excessive queries. Secondly, unlike the standard box IoU, which struggles with oriented objects (\eg curved lanes) or highly parallel lanes, the proposed attention score across the lane groups can effectively capture subtle differences among these scenarios, showcasing its remarkable flexibility.
\subsection{Loss Function}
The proportion of lane markings in an image is relatively small, especially for the curved centers. Consequently, during the training process, there exists an issue of imbalanced distribution between positive (\ie points near seed points) and negative samples (\ie no lane points). To address this problem, we employ a weighted version of focal loss \cite{Alpher26} for the centerness branch:
\begin{equation}
{L}_{ctr}= 
     \begin{cases} 
     -{(1-p)}^{\alpha}\log_{}{p}, & y >= t_{ctr}\\
     -{|1-y|^\beta}{p}^{\alpha}\log_{}{(1-p)}, & y < t_{ctr} 
     \end{cases}   
\label{eq:focalloss}
\end{equation}
where $y$ denotes the ground-truth centerness of the sample point, and $p$ represents the predicted scores. $\alpha$ serves to adjust the relative weights between high-peak center points and the background, considering the potentially significant disparity in the distribution of positive and negative samples. $\beta$ is responsible for controlling the prediction scores, increasing the weight of a sample when the predicted centerness deviates substantially from the ground-truth score. For the implementation, we set the threshold of centerness $t_{ctr}$ to 0.95, and the parameter $\alpha$ and $\beta$ is set to 2 and 4. For lane instance prediction, we utilize soft dice loss \cite{Alpher27} that is similar to \cref{eq:set_soft_iou} to supervise the seed mask, and the formula is as follows:
\begin{equation}
{L}_{inst} = \frac{1}{k} \sum_{i=0}^k 1 - \hat{IoU}(X_{i}, GT_{i}) 
\label{eq:inst_loss}
\end{equation}
where $k$ represents number of selected center points, $X_{i}$ denotes the prediction grouping mask and $GT_{i}$ is the ground-truth lane map for the corresponding seeds. We also train the semantic head as an auxiliary branch where lane group points are viewed as foreground and dice loss is utilized to train on the groud-truth mask. To sum up, the overall loss function is as follows:
\begin{equation}
L={L}_{ctr} + {L}_{inst} + {L}_{sem}
\label{eq:overall_loss}
\end{equation}

\section{Experimental Studies}
\subsection{Datasets Description}
The proposed framework is evaluated on two widely used datasets, namely TuSimple \cite{Alpher20} and CULane \cite{Alpher03}. The TuSimple dataset includes a total of 3,268 training samples and 2,782 testing samples while the CULane dataset consists of a significantly larger set, with 88,880 training images and 34,680 testing images and encompasses more diverse scenarios of nine distinct scenes (normal driving conditions, crowded scenes, and challenging lighting conditions, \etc).
\subsection{Evaluation Metrics}
We utilize $F1$ score to measure the lane detection performance:
\begin{equation}
{F}_{1}=\frac{2\cdotp Precision\cdotp Recall}{Precision+Recall}
\end{equation}
where $Precision=\frac{TP}{TP+FP}$ and $Recall=\frac{TP}{TP+FN}$, $TP$ (True Positives), $FP$ (False Positives), and $FN$ (False Negatives) represent the number of correctly identified positive samples, the number of falsely identified negative samples, and the number of missed positive samples, respectively. 
For CULane \cite{Alpher03}, a $TP$ is set to have that over 0.5 IoU \ref{eq:set_iou} with the ground-truth. For TuSimple benchmark, IoU is re-defined as an accuracy term as follows:  
\begin{equation}
accuracy=\frac{\textstyle\sum_{clip}^{}{C}_{clip}}{\textstyle\sum_{clip}^{}{S}_{clip}}
\end{equation}
where \textit{S\textsubscript{clip}} represents the total number of points in the ground truth and \textit{C\textsubscript{clip}} denotes the count of points predicted by the model, predicted lanes are considered correct when the distance between detected points and the ground-truth is less than 20 pixels and the total amount of such match exceeds 85$\%$.
\subsection{Implementation Details}
We choose HRNet \cite{Alpher23} as the convolutional backbone and utilize the Adam optimizer \cite{adam_opt} with an initial learning rate of 0.01. We decay the learning rate by 0.1 once reaches $60\%$ and $85\%$ of the total epochs. We plan to train the model for 30 epochs on the Tusimple dataset and 10 epochs on the CULane dataset. We employ data augmentation techniques (\ie random horizontal flipping, random brightness, and contrast adjustments). When training TuSimple, the number of seed points is set to 5, while for CULane, it is set to 4. During testing, seed points are uniformly configured to be 5. The batch size will be set to 16 for all training sessions on a single Tesla V100 GPU.
\subsection{Ablation Study}
In this section, we conducted ablation experiments on the TuSimple dataset to validate the effectiveness of the proposed center voting and lane grouping modules.
\begin{table}[]
\centering
\begin{tabular}{ccccc}
\hline
Width             & Aggregation & Acc   & FP   & FN   \\ \hline
 18 &             & 95.26 & 4.26 & 3.07 \\
 18 & \checkmark             & 95.80 & 4.54 & 3.01 \\
 34 &             & 95.94 & 4.55 & 2.79 \\
 34 & \checkmark            & 95.84 & 3.96 & 2.83 \\
 48 &             & 96.03 & 4.40 & 2.57 \\ 
 48 & \checkmark             & 96.10 & 3.38 & 2.39 \\ \hline
\end{tabular}
  \caption{Comparison of model performance on the TuSimple dataset with different backbone designs. Width denotes different backbone sizes and higher values contribute to larger model capacity. The second column indicates whether we aggregate backbone features from multi-scales.}
  \label{different backbone versions}
\end{table}
\begin{table}[]
\begin{tabular}{cccccc}
\hline
Centerness Type & Semantic & Acc & FP & FN \\ \hline
B    &          & 94.77   & 4.54    &  5.01     \\
C    &          & 95.13    & 6.06   & 3.55    \\ 
C    & \checkmark        & 95.26   & 4.26   & 3.07     \\ \hline
\end{tabular}
  \caption{In the first column, B indicates the box centerness using \cref{eq:fcos}, while C denotes the proposed curve-aware version using \cref{eq:changeSi}. The second column denotes whether models are trained with an auxiliary semantic head.}
  \label{ablation}
\end{table}

\noindent
\textbf{Choice of Backbones}\label{sec_exp_backbone}
We compare the performance of models with different backbone versions (\ie HRNet-18, HRNet-34 and HRNet-48) on the TuSimple dataset employed. It is observed that HRNet-48 and HRNet-34 exhibit accuracy improvements of 0.77 and 0.68 when compared to HRNet-18 baselines. We also explored feature aggregation for different backbone versions, and the corresponding experimental results in \cref{different backbone versions} show the accuracy of HRNet-48 experiences a slight decrease of 0.1, while HRNet-18 and HRNet-48 show accuracy improvements of 0.54 and 0.07 respectively.\\
\noindent
\textbf{Point Centerness Matters} 
Observing the first two rows of \cref{ablation}, it is evident that the curve path integration method proves effective in lane detection. This technique assists the model in identifying a greater number of lane lines, resulting in a reduction of the FN metric by 1.46. Consequently, this further enhances the accuracy by 0.36.\\
\noindent
\textbf{Auxiliary Semantic Branch Helps}
Comparing the last two rows of \cref{ablation}, we observe that the auxiliary semantic branch enhances the detection performance of the model. The global semantic information it provides contributes to a reduction in the model's false positive rate, resulting in a decrease of 0.2 in FP while concurrently reducing FN by 0.48, ultimately leading to a notable improvement in accuracy by 0.13.\\
\noindent
\textbf{Few-Shot Prediction via C-FPS}
For the TuSimple benchmark, each frame is annotated to have at most 5 lanes per image. As LanePtrNet can be modulated with the number of sampled points and easily output the dynamic number of detected lanes, we observe the recall and precision metrics versus the number of sampled seeds to see how the overall trend. With one-shot prediction, we have surprisingly 1.33 FP and reached 3.32 FN when we have five-shot which is already equivalent to the maximum number of lane lines per frame. Consequently, our model can switch to different modes for either higher recall or precision by simply tuning the hyper-parameter of C-FPS.

\begin{table}[]
\centering
\begin{tabular}{ccccccc}
\hline
Seed Number & 1     & 2     & 5     & 10    \\ \hline
FN          & 70.14 & 41.12  & 3.32 & 3.08 \\ 
FP       & 1.33 & 2.71 & 3.81 & 4.21 \\ \hline
\end{tabular}
    \caption{The recall and precision metrics vary with changes in the number of seeds.}
    \label{fig:recall}
\end{table}
\begin{figure*}
    \centering
    \includegraphics[width=1\linewidth]{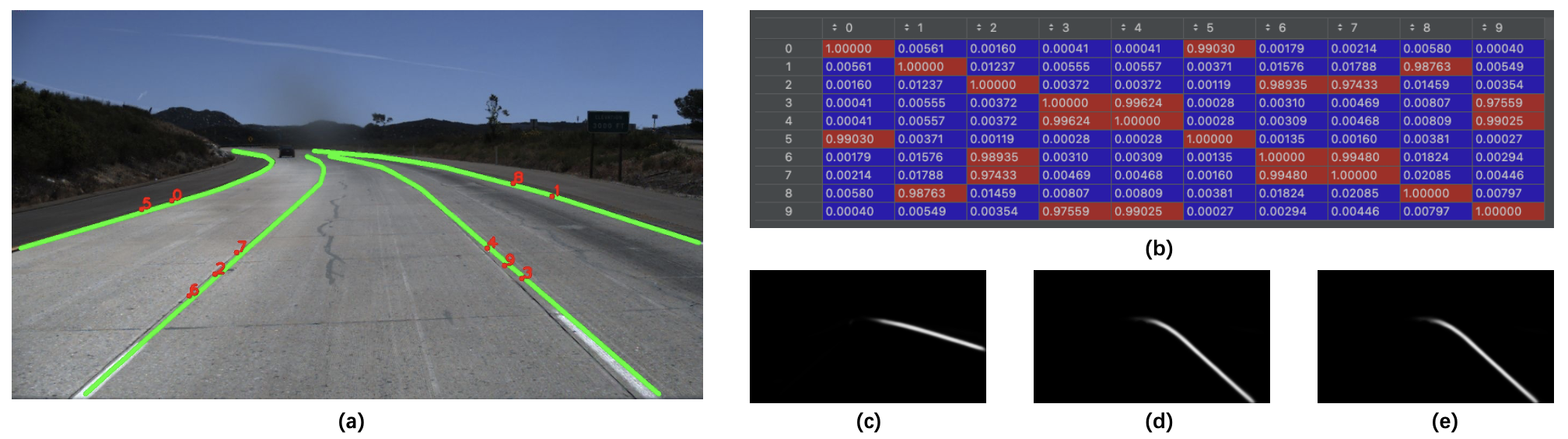}
    \caption{(a) shows predicted seed points ordered in numbers from 0 to 9. Green lines are ground-truth results for this case. (b) denotes attention scores across lane groups given centers. Red regions indicate highly correlated groups while blue blocks show unrelated prediction. (c)(d)(e) indicate the 1st,3rd and 4th lane results for reference.}
    \label{fig:attn_score}
\end{figure*}
\noindent
\textbf{Effect of Cross-instance Attention}
\cref{fig:attn_score} presents a qualitative result obtained from the test set of the TuSimple benchmark. In this particular example, we employ the C-FPS algorithm to predict 10 seed points for lane detection. These seed points are strategically positioned near the center of each lane, ensuring accurate representation while some certain lanes may exhibit repeated seed votes. Moreover, we observe the $10\times{10}$ attention map to verify the intra and inter-relationship between lane groups. It is observed that duplicate votes tend to receive remarkably high scores. Conversely, when encountering two distinct lane lines, the similarity score drops significantly to a near-zero score. This demonstrates the effectiveness of the cross-instance attention voting process in reducing the impact of duplicates and enhancing the overall accuracy.
\begin{table}[]
\centering
\begin{tabular}{cccc}
\hline
Method                  & Acc   & FP   & FN   \\ \hline
RESA \cite{Alpher04}            & 96.82 & 3.63 & 2.48 \\
CondLaneNet \cite{Alpher15}    & 96.54 & \textbf{2.01} & 3.50 \\
LaneATT \cite{Alpher08}        & 95.63 & 3.53 & 2.92 \\
CLRNet \cite{Alpher09}         & 96.83 & 2.37 & 2.38 \\
ADNet \cite{Alpher19}         & 96.60 & 2.83 & 2.53 \\
BézierLaneNet \cite{Alpher12} & 95.65 & 5.10 & 3.90 \\
FOLOLane \cite{Alpher02}      & \textbf{96.92} & 4.47 & 2.28 \\
GANet \cite{Alpher14}         & 96.44 & 2.63 & 2.47 \\
RCLane \cite{Alpher13}        & 96.58 & 2.28 & \textbf{2.27} \\ \hline
LanePtrNet              & 96.10 & 3.38 & 2.39 \\ \hline
\end{tabular}
  \caption{Performance of different methods on TuSimple.}
  \label{tusimpletab}
\end{table}
\begin{table*}[]
\begin{tabular}{llllllllllll}
\hline
Method                    & Total & Normal & Crowded & Dazzle & Shadow & No line & Arrow & Curve & Cross & Night  \\ \hline
RESA \cite{Alpher04}             & 75.30 & 92.10  & 73.10   & 69.20  & 72.80  & 47.70   & 88.30 & 70.30 & 1503  & 69.90    \\
CondLaneNet \cite{Alpher15}   & 79.48 & 93.47  & 77.44   & 70.93  & 80.91  & 54.13   & 90.16 & 75.21 & 1201  & 74.80    \\
Chen \cite{Alpher16}            & \textbf{80.77} & \textbf{94.17}  & \textbf{79.90}   & \textbf{75.43}  & 80.99  & 55.00   & \textbf{90.97} & 76.87 & 1047  & 75.11    \\
LaneATT \cite{Alpher08}           & 77.02 & 91.74  & 76.16   & 69.47  & 76.31  & 50.46   & 86.29 & 64.05 & 1264  & 70.81    \\
CLRNet \cite{Alpher09}            & 80.47 & 93.73  & 79.59   & 75.30  & \textbf{82.51}  & 54.58   & 90.62 & 74.13 & 1155  & \textbf{75.37}    \\
ADNet \cite{Alpher19}            & 78.94 & 92.90  & 77.45   & 71.71  & 79.11  & 52.89   & 89.90 & 70.64 & 1499  & 74.78    \\
BézierLaneNet   \cite{Alpher12} & 75.57 & 91.59  & 73.20   & 69.20  & 76.74  & 48.05   & 87.16 & 62.45 &  \textbf{888}   & 69.90     \\
FOLOLane \cite{Alpher02}         & 78.80 & 92.70  & 77.80   & 75.20  & 79.30  & 52.10   & 89.00 & 69.40 & 1569  & 74.50    \\
GANet \cite{Alpher14}            & 79.63 & 93.67  & 78.66   & 71.82  & 78.32  & 53.38   & 89.86 & 77.37 & 1352  & 73.85    \\
RCLane \cite{Alpher13}           & 80.50 & 94.01  & 79.13   & 72.92  & 81.16  & 53.94   & 90.51 & \textbf{79.50} &  931   & 75.10  \\ \hline
LanePtrNet                & 79.82       & 92.02       & 79.88        & 67.77       & 81.95    & \textbf{58.86}         & 87.59      & 66.61      & 1574      & 75.35           \\ \hline
\end{tabular}
  \caption{Comparison with state-of-the-art methods on CULane test set. The evaluation metric is F1 score with IoU threshold=0.5. For cross scenario, only FP are shown.}
  \label{culanetab}
\end{table*}
\subsection{Comparison with Baseline Approaches}
\noindent
\textbf{TuSimple Benchmark}
As presented in \cref{tusimpletab} LanePtrNet achieves a competitive accuracy of 96.10, demonstrating its effectiveness in lane detection. The proposed method exhibits a favorable balance between accuracy and FP metric, with a relatively low FP of 3.38. This is notably better than RESA, FOLOLane and BézierLaneNet, which have higher FP at 3.63, 4.47 and 5.10, respectively. Furthermore, LanePtrNet showcases a competitive performance in mitigating FN, with a rate of 2.39, surpassing the top-performing methods including LaneATT, BézierLaneNet, CondLaneNet, ADNet and GANet.\\
\noindent
\textbf{CULane Benchmark}
All experiments \cref{culanetab} on the CULane dataset were conducted using HRNet-18, whereas other models (\ie RCLane, CLRNet, and CondLaneNet) have computionally heavier backbones compared to our baseline. Increasing the backbone width could further enhance the total F1 score, as demonstrated in \cref{sec_exp_backbone}. In the total F1 rankings, LanePtrNet claims the fourth position, indicating overall effectiveness in lane detection. In the no-line scenario, LanePtrNet exhibits superior performance, ranking first among other models. In night scenes, LanePtrNet secures the second position, trailing only 0.02 behind CLRNet. In crowded scenarios, it also holds the second spot, with a marginal 0.02 difference compared to \cite{Alpher16}.

In summary, LanePtrNet stands out as a robust lane detection model on the both datasets, excelling in accuracy while maintaining a favorable balance between FP and FN metrics. This highlights its potential as a reliable solution for practical applications in autonomous driving and computer vision tasks.\\
\section{Conclusion}
In this paper, we introduce a novel framework that leverages seed point voting and grouping on curves to accurately detect lane lines. We design a curve-aware centerness as a crucial measurement and propose the C-FPS algorithm to extract seed points. To generate lane clusters, we propose a simple yet effective grouping module and adopt cross-instance attention voting that helps to significantly reduce duplicate grouping results. Experimental results demonstrate the competitive performance of LanePtrNet compared to other baseline approaches.

In future work, we plan to address the following aspects: Firstly, we aim to improve the loss function used in the seed voting mechanism, as in \cite{gfocal_loss}\cite{gfocal_loss_v2}, a unified quality score proves feasible. Secondly, we plan to optimize the grouping module by incorporating multi-scale feature aggregation instead of operating at a single scale. Furthermore, We intend to pursue further research on 3D lane detection applications. Since our framework is based on an encoder-decoder architecture and can be easily transformed into a completely point-based model. This allows us to handle data from various modalities (\ie camera, lidar, radar, IMU data, \etc), all of which can be represented as data points in a set.
{
    \small
    \bibliographystyle{ieeenat_fullname}

\begin{thebibliography}{48}
\providecommand{\natexlab}[1]{#1}
\providecommand{\url}[1]{\texttt{#1}}
\expandafter\ifx\csname urlstyle\endcsname\relax
  \providecommand{\doi}[1]{doi: #1}\else
  \providecommand{\doi}{doi: \begingroup \urlstyle{rm}\Url}\fi

\bibitem[Behrendt and Soussan(2019)]{Alpher21}
Karsten Behrendt and Ryan Soussan.
\newblock Unsupervised labeled lane markers using maps.
\newblock In \emph{Proceedings of the IEEE/CVF international conference on computer vision workshops}, pages 0--0, 2019.

\bibitem[Cai et~al.(2020)Cai, Wang, Luo, Yin, Du, Wang, Zhang, Zhou, Zhou, and Sun]{pose_rsn}
Yuanhao Cai, Zhicheng Wang, Zhengxiong Luo, Binyi Yin, Angang Du, Haoqian Wang, Xiangyu Zhang, Xinyu Zhou, Erjin Zhou, and Jian Sun.
\newblock Learning delicate local representations for multi-person pose estimation.
\newblock In \emph{Computer Vision--ECCV 2020: 16th European Conference, Glasgow, UK, August 23--28, 2020, Proceedings, Part III 16}, pages 455--472. Springer, 2020.

\bibitem[Chen et~al.(2022)Chen, Chen, Zhang, and Tao]{sasa-fps}
Chen Chen, Zhe Chen, Jing Zhang, and Dacheng Tao.
\newblock Sasa: Semantics-augmented set abstraction for point-based 3d object detection.
\newblock In \emph{Proceedings of the AAAI Conference on Artificial Intelligence}, pages 221--229, 2022.

\bibitem[Chen et~al.(2017)Chen, Papandreou, Kokkinos, Murphy, and Yuille]{chen2017deeplab}
Liang-Chieh Chen, George Papandreou, Iasonas Kokkinos, Kevin Murphy, and Alan~L Yuille.
\newblock Deeplab: Semantic image segmentation with deep convolutional nets, atrous convolution, and fully connected crfs.
\newblock \emph{IEEE transactions on pattern analysis and machine intelligence}, 40\penalty0 (4):\penalty0 834--848, 2017.

\bibitem[Chen et~al.(2023)Chen, Liu, Gong, Du, Qian, and Smith-Miles]{Alpher16}
Ziye Chen, Yu Liu, Mingming Gong, Bo Du, Guoqi Qian, and Kate Smith-Miles.
\newblock Generating dynamic kernels via transformers for lane detection.
\newblock In \emph{Proceedings of the IEEE/CVF International Conference on Computer Vision}, pages 6835--6844, 2023.

\bibitem[Choy et~al.(2019)Choy, Gwak, and Savarese]{minkunet}
Christopher Choy, JunYoung Gwak, and Silvio Savarese.
\newblock 4d spatio-temporal convnets: Minkowski convolutional neural networks.
\newblock In \emph{Proceedings of the IEEE/CVF conference on computer vision and pattern recognition}, pages 3075--3084, 2019.

\bibitem[Dai et~al.(2016)Dai, Li, He, and Sun]{rfcn}
Jifeng Dai, Yi Li, Kaiming He, and Jian Sun.
\newblock R-fcn: Object detection via region-based fully convolutional networks.
\newblock \emph{Advances in neural information processing systems}, 29, 2016.

\bibitem[Feng et~al.(2022)Feng, Guo, Tan, Xu, Wang, and Ma]{Alpher12}
Zhengyang Feng, Shaohua Guo, Xin Tan, Ke Xu, Min Wang, and Lizhuang Ma.
\newblock Rethinking efficient lane detection via curve modeling.
\newblock In \emph{Proceedings of the IEEE/CVF Conference on Computer Vision and Pattern Recognition}, pages 17062--17070, 2022.

\bibitem[Haotian et~al.(2023)Haotian, Fanyi, Jingwen, Shiyu, and Zhiwang]{ic-fps}
Hu Haotian, Wang Fanyi, Su Jingwen, Gao Shiyu, and Zhang Zhiwang.
\newblock Ic-fps: Instance-centroid faster point sampling module for 3d point-base object detection.
\newblock \emph{arXiv preprint arXiv:2303.17921}, 2023.

\bibitem[He et~al.(2016)He, Zhang, Ren, and Sun]{resnet}
Kaiming He, Xiangyu Zhang, Shaoqing Ren, and Jian Sun.
\newblock Deep residual learning for image recognition.
\newblock In \emph{Proceedings of the IEEE conference on computer vision and pattern recognition}, pages 770--778, 2016.

\bibitem[He et~al.(2017)He, Gkioxari, Doll{\'a}r, and Girshick]{maskrcnn}
Kaiming He, Georgia Gkioxari, Piotr Doll{\'a}r, and Ross Girshick.
\newblock Mask r-cnn.
\newblock In \emph{Proceedings of the IEEE international conference on computer vision}, pages 2961--2969, 2017.

\bibitem[Honda and Uchida(2023)]{Alpher07}
Hiroto Honda and Yusuke Uchida.
\newblock Clrernet: Improving confidence of lane detection with laneiou.
\newblock \emph{arXiv preprint arXiv:2305.08366}, 2023.

\bibitem[Ioffe and Szegedy(2015)]{batchnorm}
Sergey Ioffe and Christian Szegedy.
\newblock Batch normalization: Accelerating deep network training by reducing internal covariate shift.
\newblock In \emph{International conference on machine learning}, pages 448--456. pmlr, 2015.

\bibitem[Kingma and Ba(2014)]{adam_opt}
Diederik~P Kingma and Jimmy Ba.
\newblock Adam: A method for stochastic optimization.
\newblock \emph{arXiv preprint arXiv:1412.6980}, 2014.

\bibitem[Ko et~al.(2021)Ko, Lee, Azam, Munir, Jeon, and Pedrycz]{Alpher17}
Yeongmin Ko, Younkwan Lee, Shoaib Azam, Farzeen Munir, Moongu Jeon, and Witold Pedrycz.
\newblock Key points estimation and point instance segmentation approach for lane detection.
\newblock \emph{IEEE Transactions on Intelligent Transportation Systems}, 23\penalty0 (7):\penalty0 8949--8958, 2021.

\bibitem[Li et~al.(2020)Li, Wang, Wu, Chen, Hu, Li, Tang, and Yang]{gfocal_loss}
Xiang Li, Wenhai Wang, Lijun Wu, Shuo Chen, Xiaolin Hu, Jun Li, Jinhui Tang, and Jian Yang.
\newblock Generalized focal loss: Learning qualified and distributed bounding boxes for dense object detection.
\newblock \emph{Advances in Neural Information Processing Systems}, 33:\penalty0 21002--21012, 2020.

\bibitem[Li et~al.(2021{\natexlab{a}})Li, Wang, Hu, Li, Tang, and Yang]{gfocal_loss_v2}
Xiang Li, Wenhai Wang, Xiaolin Hu, Jun Li, Jinhui Tang, and Jian Yang.
\newblock Generalized focal loss v2: Learning reliable localization quality estimation for dense object detection.
\newblock In \emph{Proceedings of the IEEE/CVF Conference on Computer Vision and Pattern Recognition}, pages 11632--11641, 2021{\natexlab{a}}.

\bibitem[Li et~al.(2021{\natexlab{b}})Li, Hou, Wu, Jiao, Ren, and Yang]{li2021fcosr}
Zhonghua Li, Biao Hou, Zitong Wu, Licheng Jiao, Bo Ren, and Chen Yang.
\newblock Fcosr: A simple anchor-free rotated detector for aerial object detection.
\newblock \emph{arXiv preprint arXiv:2111.10780}, 2021{\natexlab{b}}.

\bibitem[Lin et~al.(2017)Lin, Goyal, Girshick, He, and Doll{\'a}r]{Alpher26}
Tsung-Yi Lin, Priya Goyal, Ross Girshick, Kaiming He, and Piotr Doll{\'a}r.
\newblock Focal loss for dense object detection.
\newblock In \emph{Proceedings of the IEEE international conference on computer vision}, pages 2980--2988, 2017.

\bibitem[Liu et~al.(2021{\natexlab{a}})Liu, Chen, Zhu, and Tan]{Alpher15}
Lizhe Liu, Xiaohao Chen, Siyu Zhu, and Ping Tan.
\newblock Condlanenet: a top-to-down lane detection framework based on conditional convolution.
\newblock In \emph{Proceedings of the IEEE/CVF International Conference on Computer Vision}, pages 3773--3782, 2021{\natexlab{a}}.

\bibitem[Liu et~al.(2021{\natexlab{b}})Liu, Yuan, Liu, and Xiong]{Alpher11}
Ruijin Liu, Zejian Yuan, Tie Liu, and Zhiliang Xiong.
\newblock End-to-end lane shape prediction with transformers.
\newblock In \emph{Proceedings of the IEEE/CVF winter conference on applications of computer vision}, pages 3694--3702, 2021{\natexlab{b}}.

\bibitem[Liu et~al.(2019)Liu, Tang, Lin, and Han]{spvcnn}
Zhijian Liu, Haotian Tang, Yujun Lin, and Song Han.
\newblock Point-voxel cnn for efficient 3d deep learning.
\newblock \emph{Advances in Neural Information Processing Systems}, 32, 2019.

\bibitem[Long et~al.(2015)Long, Shelhamer, and Darrell]{fcn_seg}
Jonathan Long, Evan Shelhamer, and Trevor Darrell.
\newblock Fully convolutional networks for semantic segmentation.
\newblock In \emph{Proceedings of the IEEE conference on computer vision and pattern recognition}, pages 3431--3440, 2015.

\bibitem[Milletari et~al.(2016)Milletari, Navab, and Ahmadi]{Alpher27}
Fausto Milletari, Nassir Navab, and Seyed-Ahmad Ahmadi.
\newblock V-net: Fully convolutional neural networks for volumetric medical image segmentation.
\newblock In \emph{2016 fourth international conference on 3D vision (3DV)}, pages 565--571. Ieee, 2016.

\bibitem[Moon et~al.(2019)Moon, Chang, and Lee]{moon2019posefix}
Gyeongsik Moon, Ju~Yong Chang, and Kyoung~Mu Lee.
\newblock Posefix: Model-agnostic general human pose refinement network.
\newblock In \emph{Proceedings of the IEEE/CVF Conference on Computer Vision and Pattern Recognition}, pages 7773--7781, 2019.

\bibitem[Pan et~al.(2018)Pan, Shi, Luo, Wang, and Tang]{Alpher03}
Xingang Pan, Jianping Shi, Ping Luo, Xiaogang Wang, and Xiaoou Tang.
\newblock Spatial as deep: Spatial cnn for traffic scene understanding.
\newblock In \emph{Proceedings of the AAAI Conference on Artificial Intelligence}, 2018.

\bibitem[Qi et~al.(2017{\natexlab{a}})Qi, Su, Mo, and Guibas]{pointnet}
Charles~R Qi, Hao Su, Kaichun Mo, and Leonidas~J Guibas.
\newblock Pointnet: Deep learning on point sets for 3d classification and segmentation.
\newblock In \emph{Proceedings of the IEEE conference on computer vision and pattern recognition}, pages 652--660, 2017{\natexlab{a}}.

\bibitem[Qi et~al.(2017{\natexlab{b}})Qi, Yi, Su, and Guibas]{Alpher25}
Charles~Ruizhongtai Qi, Li Yi, Hao Su, and Leonidas~J Guibas.
\newblock Pointnet++: Deep hierarchical feature learning on point sets in a metric space.
\newblock \emph{Advances in neural information processing systems}, 30, 2017{\natexlab{b}}.

\bibitem[Qin et~al.(2020)Qin, Wang, and Li]{Alpher06}
Zequn Qin, Huanyu Wang, and Xi Li.
\newblock Ultra fast structure-aware deep lane detection.
\newblock In \emph{Computer Vision--ECCV 2020: 16th European Conference, Glasgow, UK, August 23--28, 2020, Proceedings, Part XXIV 16}, pages 276--291. Springer, 2020.

\bibitem[Qu et~al.(2021)Qu, Jin, Zhou, Yang, and Zhang]{Alpher02}
Zhan Qu, Huan Jin, Yang Zhou, Zhen Yang, and Wei Zhang.
\newblock Focus on local: Detecting lane marker from bottom up via key point.
\newblock In \emph{Proceedings of the IEEE/CVF Conference on Computer Vision and Pattern Recognition}, pages 14122--14130, 2021.

\bibitem[Shi et~al.(2020)Shi, Guo, Jiang, Wang, Shi, Wang, and Li]{pvrcnn}
Shaoshuai Shi, Chaoxu Guo, Li Jiang, Zhe Wang, Jianping Shi, Xiaogang Wang, and Hongsheng Li.
\newblock Pv-rcnn: Point-voxel feature set abstraction for 3d object detection.
\newblock In \emph{Proceedings of the IEEE/CVF conference on computer vision and pattern recognition}, pages 10529--10538, 2020.

\bibitem[Tabelini et~al.(2021{\natexlab{a}})Tabelini, Berriel, Paixao, Badue, De~Souza, and Oliveira-Santos]{Alpher08}
Lucas Tabelini, Rodrigo Berriel, Thiago~M Paixao, Claudine Badue, Alberto~F De~Souza, and Thiago Oliveira-Santos.
\newblock Keep your eyes on the lane: Real-time attention-guided lane detection.
\newblock In \emph{Proceedings of the IEEE/CVF conference on computer vision and pattern recognition}, pages 294--302, 2021{\natexlab{a}}.

\bibitem[Tabelini et~al.(2021{\natexlab{b}})Tabelini, Berriel, Paixao, Badue, De~Souza, and Oliveira-Santos]{Alpher10}
Lucas Tabelini, Rodrigo Berriel, Thiago~M Paixao, Claudine Badue, Alberto~F De~Souza, and Thiago Oliveira-Santos.
\newblock Polylanenet: Lane estimation via deep polynomial regression.
\newblock In \emph{2020 25th International Conference on Pattern Recognition (ICPR)}, pages 6150--6156. IEEE, 2021{\natexlab{b}}.

\bibitem[{TuSimple}(2017)]{Alpher20}
{TuSimple}.
\newblock Tusimple lane detection benchmark.
\newblock \url{https://github.com/TuSimple/tusimple-benchmark}, 2017.

\bibitem[Van~Gansbeke et~al.(2019)Van~Gansbeke, De~Brabandere, Neven, Proesmans, and Van~Gool]{Alpher22}
Wouter Van~Gansbeke, Bert De~Brabandere, Davy Neven, Marc Proesmans, and Luc Van~Gool.
\newblock End-to-end lane detection through differentiable least-squares fitting.
\newblock In \emph{Proceedings of the IEEE/CVF International Conference on Computer Vision Workshops}, pages 0--0, 2019.

\bibitem[Wang et~al.(2020)Wang, Sun, Cheng, Jiang, Deng, Zhao, Liu, Mu, Tan, Wang, et~al.]{Alpher23}
Jingdong Wang, Ke Sun, Tianheng Cheng, Borui Jiang, Chaorui Deng, Yang Zhao, Dong Liu, Yadong Mu, Mingkui Tan, Xinggang Wang, et~al.
\newblock Deep high-resolution representation learning for visual recognition.
\newblock \emph{IEEE transactions on pattern analysis and machine intelligence}, 43\penalty0 (10):\penalty0 3349--3364, 2020.

\bibitem[Wang et~al.(2021)Wang, Sun, Cheng, Jiang, Deng, Zhao, Liu, Mu, Tan, Wang, Liu, and Xiao]{Alpher24}
Jingdong Wang, Ke Sun, Tianheng Cheng, Borui Jiang, Chaorui Deng, Yang Zhao, Dong Liu, Yadong Mu, Mingkui Tan, Xinggang Wang, Wenyu Liu, and Bin Xiao.
\newblock Deep high-resolution representation learning for visual recognition.
\newblock \emph{IEEE Transactions on Pattern Analysis and Machine Intelligence}, 43\penalty0 (10):\penalty0 3349--3364, 2021.

\bibitem[Wang et~al.(2022)Wang, Ma, Huang, Hui, Wang, Qian, and Zhang]{Alpher14}
Jinsheng Wang, Yinchao Ma, Shaofei Huang, Tianrui Hui, Fei Wang, Chen Qian, and Tianzhu Zhang.
\newblock A keypoint-based global association network for lane detection.
\newblock In \emph{Proceedings of the IEEE/CVF Conference on Computer Vision and Pattern Recognition}, pages 1392--1401, 2022.

\bibitem[Xiao et~al.(2023)Xiao, Li, Yang, and Yang]{Alpher19}
Lingyu Xiao, Xiang Li, Sen Yang, and Wankou Yang.
\newblock Adnet: Lane shape prediction via anchor decomposition.
\newblock In \emph{Proceedings of the IEEE/CVF International Conference on Computer Vision}, pages 6404--6413, 2023.

\bibitem[Xie et~al.(2021)Xie, Cheng, Wang, Yao, and Han]{orcnn}
Xingxing Xie, Gong Cheng, Jiabao Wang, Xiwen Yao, and Junwei Han.
\newblock Oriented r-cnn for object detection.
\newblock In \emph{Proceedings of the IEEE/CVF international conference on computer vision}, pages 3520--3529, 2021.

\bibitem[Xu et~al.(2022)Xu, Cai, Zhao, Zhang, Xu, Fu, and Xue]{Alpher13}
Shenghua Xu, Xinyue Cai, Bin Zhao, Li Zhang, Hang Xu, Yanwei Fu, and Xiangyang Xue.
\newblock Rclane: Relay chain prediction for lane detection.
\newblock In \emph{European Conference on Computer Vision}, pages 461--477. Springer, 2022.

\bibitem[Zhang et~al.(2020)Zhang, Chi, Yao, Lei, and Li]{atss}
Shifeng Zhang, Cheng Chi, Yongqiang Yao, Zhen Lei, and Stan~Z Li.
\newblock Bridging the gap between anchor-based and anchor-free detection via adaptive training sample selection.
\newblock In \emph{Proceedings of the IEEE/CVF conference on computer vision and pattern recognition}, pages 9759--9768, 2020.

\bibitem[Zhang et~al.(2021)Zhang, Zhu, Feng, Fu, Wang, Li, Li, and Wang]{Alpher05}
Yujun Zhang, Lei Zhu, Wei Feng, Huazhu Fu, Mingqian Wang, Qingxia Li, Cheng Li, and Song Wang.
\newblock Vil-100: A new dataset and a baseline model for video instance lane detection.
\newblock In \emph{Proceedings of the IEEE/CVF International Conference on Computer Vision}, pages 15681--15690, 2021.

\bibitem[Zheng et~al.(2021)Zheng, Fang, Zhang, Tang, Yang, Liu, and Cai]{Alpher04}
Tu Zheng, Hao Fang, Yi Zhang, Wenjian Tang, Zheng Yang, Haifeng Liu, and Deng Cai.
\newblock Resa: Recurrent feature-shift aggregator for lane detection.
\newblock In \emph{Proceedings of the AAAI Conference on Artificial Intelligence}, pages 3547--3554, 2021.

\bibitem[Zheng et~al.(2022)Zheng, Huang, Liu, Tang, Yang, Cai, and He]{Alpher09}
Tu Zheng, Yifei Huang, Yang Liu, Wenjian Tang, Zheng Yang, Deng Cai, and Xiaofei He.
\newblock Clrnet: Cross layer refinement network for lane detection.
\newblock In \emph{Proceedings of the IEEE/CVF conference on computer vision and pattern recognition}, pages 898--907, 2022.

\bibitem[Zhou et~al.(2019)Zhou, Wang, and Kr{\"a}henb{\"u}hl]{centernet}
Xingyi Zhou, Dequan Wang, and Philipp Kr{\"a}henb{\"u}hl.
\newblock Objects as points.
\newblock \emph{arXiv preprint arXiv:1904.07850}, 2019.

\bibitem[Zhou and Tuzel(2018)]{voxelnet}
Yin Zhou and Oncel Tuzel.
\newblock Voxelnet: End-to-end learning for point cloud based 3d object detection.
\newblock In \emph{Proceedings of the IEEE conference on computer vision and pattern recognition}, pages 4490--4499, 2018.

\bibitem[Zou et~al.(2020)Zou, Jiang, Dai, Yue, Chen, and Wang]{Alpher01}
Qin Zou, Hanwen Jiang, Qiyu Dai, Yuanhao Yue, Long Chen, and Qian Wang.
\newblock Robust lane detection from continuous driving scenes using deep neural networks.
\newblock \emph{IEEE Transactions on Vehicular Technology}, 69\penalty0 (1):\penalty0 41--54, 2020.

\end{thebibliography}

}


\end{document}